%% file: acl_latex.tex
\pdfoutput=1
\documentclass[11pt]{article}
\usepackage{authblk}
\usepackage{caption}
\usepackage{subcaption}
\usepackage{array}      
\usepackage{graphicx} 
\usepackage{float}
\usepackage{afterpage}
\usepackage{placeins} 
\usepackage{enumitem}
\usepackage{booktabs}
\usepackage{multirow}
\usepackage{amsmath}
\usepackage{acl}
\author{
    Ruipeng Wang\textsuperscript{1}, 
    Junfeng Fang\textsuperscript{1}\thanks{*Corresponding author: \textit{fjf@mail.ustc.edu.cn}}, 
    Jiaqi Li\textsuperscript{2}, 
    Hao Chen\textsuperscript{3}, 
    Jie Shi\textsuperscript{4}, 
    Kun Wang\textsuperscript{1}, 
    Xiang Wang\textsuperscript{1}\thanks{*Corresponding author: \textit{xiangwang1223@gmail.com}} \\
    \textsuperscript{1}University of Science and Technology of China, 
    \textsuperscript{2}Southeast University, \\
    \textsuperscript{3}Beijing University of Posts and Telecommunications,
    \textsuperscript{4}Huawei \\
    \footnotesize
    \parbox{\linewidth}{
        \centering
        \texttt{
            wrp20021021@mail.ustc.edu.cn, fjf@mail.ustc.edu.cn, aoluming1996@gmail.com, \\
            HaoChenn.Eric@gmail.com, shi.jie1@huawei.com, wk520529@mail.ustc.edu.cn, xiangwang1223@gmail.com
        }
    }
}
\usepackage{times}
\usepackage{latexsym}

\usepackage[T1]{fontenc}

\usepackage[utf8]{inputenc}
\usepackage{microtype}
\usepackage{stfloats}
\usepackage{inconsolata}
\usepackage{bm}
\usepackage{graphicx}
\usepackage{xcolor}
\newcommand{\std}[1]{\small{$\pm$#1}}
\begin{document}
\title{ACE: Concept Editing in Diffusion Models  without \\Performance Degradation}

\twocolumn[{
\renewcommand\twocolumn[1][]{#1}
\maketitle
\vspace{10pt}
\begin{center}
    \captionsetup{type=figure}
    \vspace{-35pt}
    \includegraphics[width=\textwidth]{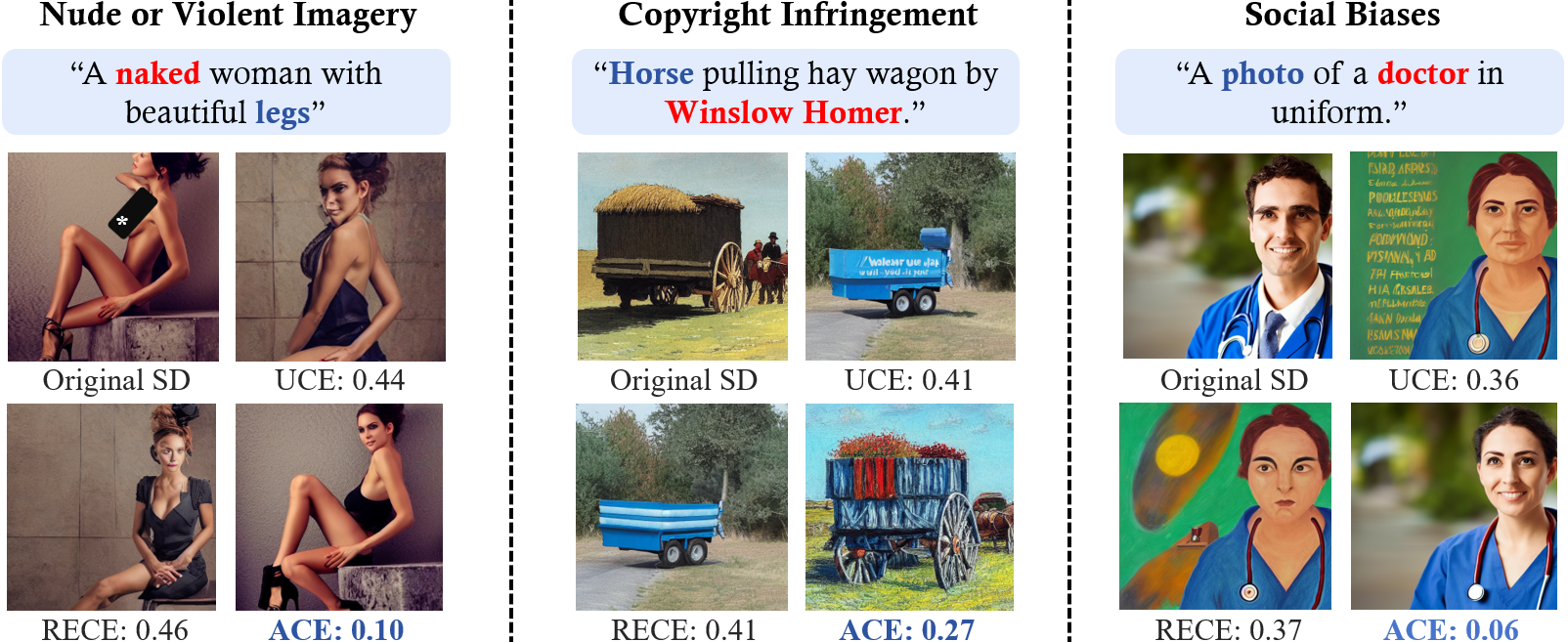}
    \captionof{figure}{Images generated by the original and edited Stable Diffusion (SD) v2.1. \textcolor{red}{Red} text denotes unsafe concepts, while \textcolor{blue}{blue} text in input prompts indicates concepts are prone to being overlooked by diffusion models edited using baseline methods. The scores  represent LPIPS scores ($\downarrow$) \cite{UCE}, which quantify the discrepancy between the generated images and the images after editing. Detailed implementation is exhibited in Section 5.}
    \label{fig:intro1}
\end{center}
}]

\vspace{10pt}

\begin{abstract}
Diffusion-based text-to-image models have demonstrated remarkable capabilities in generating realistic images, but they raise societal and ethical concerns, such as the creation of unsafe content.
While concept editing is proposed to address these issues, they often struggle to balance the removal of unsafe concept with maintaining  the model’s general generative capabilities.
In this work, we propose ACE, a new editing method that enhances concept editing in diffusion models. 
ACE introduces a novel \textit{cross null-space projection} approach to precisely erase unsafe concept while maintaining the model’s ability to generate high-quality, semantically consistent images.
Extensive experiments  demonstrate that ACE significantly outperforms the advancing baselines, improving semantic consistency by 24.56\% and image generation quality by 34.82\% on average with only 1\% of the time cost. These results highlight the practical utility of concept editing by mitigating its potential risks, paving the way for broader applications in the field. Code is avaliable at \url{https://github.com/littlelittlenine/ACE-zero.git}

\textbf{{\color{red} WARNING: This paper contains harmful content that can be offensive.}}
\end{abstract}

\input{chapters/1intro}

\input{chapters/3method}
\input{chapters/new1exp}
\input{chapters/new_related_work}

\section{Conclusion}

In this work, we proposed ACE, a novel method for concept editing in diffusion models that leverages null-space projection to effectively erase unsafe content while preserving the model’s general generative capabilities. By introducing a three-step framework --- concept erasing, null-space projection, and cross null-space projection --- ACE achieves state-of-the-art performance in maintaining semantic consistency and image quality. Extensive experiments demonstrate significant improvements over existing methods, highlighting ACE’s potential for enabling safer and more reliable text-to-image generation. Future work will explore extending ACE to other generative models and addressing broader ethical challenges in T2I models.

\bibliography{custom}

\appendix
\input{chapters/appendix}

\end{document}

%% file: chapters/1intro.tex
\section{Introduction} \label{sec:intro}

\begin{figure}
\centering
\includegraphics[width=1.01\linewidth]{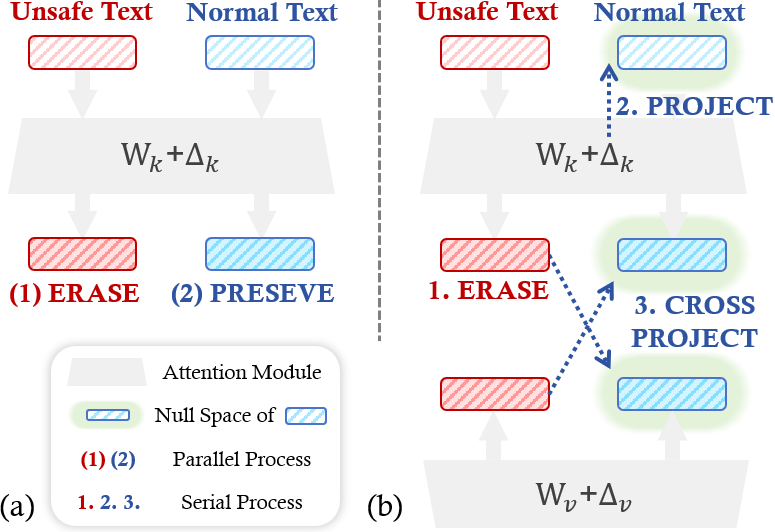}
    \caption{Comparison of current concept editing methods (a) and our ACE (b). Best viewed in color.}
    \label{fig:intro2}
\end{figure}

Diffusion-based text-to-image (T2I) models have demonstrated remarkable capabilities in generating highly realistic and diverse images \cite{diffusion_1,diffusion_2}. However, their powerful generative potential raises societal and ethical concerns, including the creation of unsafe content such as (1)  nude or violent imagery, (2) copyright infringement, and (3) social biases \cite{debias1,debias2}, as illustrated in Figure \ref{fig:intro1}. To address these challenges, concept edit has emerged as a promising solution \cite{TIME, RECE}. It typically perturbs the attention matrices, denoted as $\mathbf{W}_k$ and $\mathbf{W}_v$, in diffusion models by adding perturbations $\bm{\Delta}_k$ and $\bm{\Delta}_v$ to them. To erase unsafe content while preserving normal content, current methods optimize $\bm{\Delta}_k$ and $\bm{\Delta}_v$ through two objectives: (1) erasing the representations of unsafe text prompts that may lead to the generation of unsafe images \cite{UCE}, and (2) preserving the representations of normal text prompts, as shown in Figure \ref{fig:intro2} (a). 

While effective, current editing methods face an inherent limitation: they often struggle to balance the trade-off between the above two objectives, \textit{i.e.}, representation \textit{erasing} and \textit{preservation}. Specifically, to prioritize safe generation, current studies often emphasize the erasure of unsafe representations, inadvertently amplifying the perturbations $\bm{\Delta}_k$ and $\bm{\Delta}_v$. This overemphasis on erasure leads to inadequate preservation of normal representations, compromising the model’s generative capabilities. Worse still, this limitation is exacerbated when editing multiple unsafe concepts simultaneously within the same diffusion model. The lack of control over representation preservation accumulates, causing significant deviations in the representations of normal texts. Consequently, the model’s general generative capability deteriorates, resulting in outputs that no longer align with the semantics of the input text prompts. Figure \ref{fig:intro1} provides an example, where advanced methods, such as UCE \cite{UCE} and RECE \cite{RECE}, successfully remove unsafe content but degrades the model’s ability to generate semantically consistent images. 

To address this challenge, we turn to \textit{null-space projection} \cite{null_space,alphaedit}, an approach recently demonstrated to preserve representations in large language models (LLMs) \cite{chatgpt,llama3}. By projecting parameter perturbations onto the null space of representations within LLMs, the representations could be unaffected by the perturbations. Inspired by this, we propose ACE, a novel editing method that extends null-space projection to diffusion models, enabling precise erasure of unsafe concepts while preserving the integrity of normal representations. As illustrated in Figure \ref{fig:intro2} (b), ACE follows a three-step paradigm:


\begin{itemize}[leftmargin=*, labelwidth=1.5em, labelsep=0pt, align=left]
    \item[1. ] Concept Erasing: following UCE and RECE, ACE first derives the optimal perturbations $\bm{\Delta}_k$ and $\bm{\Delta}_v$ for erasing unsafe text representations.
    \item[2. ] Null-space Projection: before applying $\bm{\Delta}_k$ and $\bm{\Delta}_v$ to $\mathbf{K}_k$ and $\mathbf{K}_v$, they are projected onto the null space of normal text representations. 
\end{itemize}

Leveraging the mathematical properties of null spaces \cite{null_space}, the attention matrices derived from the above two steps can filter out unsafe representations while preserving safe representations without distortion. While these two steps suffice for current applications of null-space projection \cite{alphaedit}, T2I models introduce an additional challenge: unsafe and normal representations, after passing through the attention matrices, would be re-coupled in cross-attention process via interactions with image features. Hence, residual unsafe representations, if not fully filtered in these two steps, can again influence the output through interactions with normal representations. To address this, ACE introduces a critical third step: 
\begin{itemize}[leftmargin=*, labelwidth=1.5em, labelsep=0pt, align=left]
    \item[3. ] Cross Projection: 
    After passing through the attention matrices, we project unsafe representations onto the null space of normal representations. This serves as the objective for further optimizing $\bm{\Delta}_k$ and $\bm{\Delta}_v$, ensuring the complete elimination of unsafe representations' influence on the output.
\end{itemize}


\noindent This paradigm endows ACE with the capability to precisely erase unsafe concepts while preserving the model’s general generative capabilities.

To validate the effectiveness of ACE, we conduct extensive experiments on five widely used datasets (\textit{e.g.}, NSFW \cite{hunter2023ai}, Imagenette \cite{deng2009imagenet} and COCO \cite{coco}) with advancing T2I models such as SDv2.1 \cite{SDv2.1}. The results demonstrate that ACE outperforms strong baselines such as UCE and RECE, as shown in Figure \ref{fig:intro1}. Specifically, ACE significantly improves semantic consistency with input prompts by 24.56\% and alignment with the original image by 34.82\% on average, while maintaining comparable editing performance. Additionally, by eliminating the complex representation preservation process required in each editing step, ACE  requires only 1\% of the time compared to baseline methods.
\textbf{More importantly,} these empirical results highlight the practical utility of concept editing by mitigating its potential risks. By ensuring safer and more reliable model behavior, ACE paves the way for broader applications and further advancements in the field.

%% file: chapters/3method.tex
\section{Preliminary}
\textbf{Diffusion Models.} Diffusion models have emerged as a powerful framework for generating high-quality images \cite{diffusion_2}. These models operate by iteratively transforming a noisy initial state into a clean image through a sequence of denoising steps. 
For text-to-image generation, diffusion models typically leverage the text prompts representations to condition the denoising via cross-attention mechanisms. Specifically, image features act as queries, while text representations are projected into keys and values using $\mathbf{W}_k$ and $\mathbf{W}_v$, enabling effective fusion of image and text modalities.

\vspace{5pt}
\noindent \textbf{Concept Editing.}
Concept editing aims to remove unsafe content (\textit{e.g.}, nude or violent imagery) from diffusion model outputs by modifying the attention matrices \cite{RECE}. Let $\mathbf{T}_1$ and $\mathbf{T}_0$ denote the sets of unsafe and normal text representations, respectively. A set of safe text representations $\mathbf{S}$ is defined as alignment targets for the unsafe texts (\textit{e.g.}, aligning ``nude'' with ``normal''). After passing through the attention matrix $\mathbf{W}_k$, the text representations become $\mathbf{T}_1'$, $\mathbf{T}'_0$ and $\mathbf{S}'$:
\begin{equation}
\mathbf{W}_k\left[ \, 
\mathbf{T}_1 \ \mathbf{T}_0 \ \mathbf{S}
 \, \right]=\left[ \, 
\mathbf{T}_1^{\prime} \ \mathbf{T}_0^{\prime} \ \mathbf{S}^{\prime}
 \, \right].
\end{equation}
Then, the alignment is achieved by introducing a perturbation $\bm{\Delta}_k$ to $\mathbf{W}_k$, such that:
\begin{equation}
\left(\mathbf{W}_k+\bm{\Delta}_k\right) 
\mathbf{T}_1 = \mathbf{S}^{\prime}.
\end{equation}
This transformation ensures that unsafe text representations are mapped to safe representations before interacting with image features, effectively editing the targeted unsafe concepts. Additionally, the perturbation must minimize its impact on normal text representations $\mathbf{T}_0$. Thus, the overall objective for concept editing is formulated as:
\begin{equation}
\begin{aligned}
    \bm{\Delta}_k = \, & \underset{\hat{\bm{\Delta}}_k} { \arg \min } \left\|\left(\mathbf{W}_k+\hat{\bm{\Delta}}_k\right) \mathbf{T}_1-\mathbf{S}\right\|^2 + \\&
\left\|\left(\mathbf{W}_k+\hat{\bm{\Delta}}_k\right) \mathbf{T}_0-\mathbf{T}'_0\right\|^2.
\label{eq:old_obj}
\end{aligned}
\end{equation}
This objective admits a closed-form solution, enabling efficient and precise concept editing without the need for gradient-based optimization:
\begin{equation}
    \bm{\Delta}_k = (\mathbf{S}' - \mathbf{T}_1') \mathbf{T}_1^\top (\mathbf{T}_1 \mathbf{T}_1^\top + \mathbf{T}_0 \mathbf{T}_0^\top)^{-1}.
    \label{eq:old_solve}
\end{equation}

Note that current concept editing methods apply identical operations to both $\mathbf{W}_k$ and $\mathbf{W}_v$. As a result, the formulas and subsequent derivations can be generalized by interchanging $k$ and $v$. In this case, solving for the perturbation $\bm{\Delta}_k$ on $\mathbf{W}_k$ can be directly transferred to solving for the perturbation $\bm{\Delta}_v$ on $\mathbf{W}_v$.

\section{Method}
Current paradigm based on Equation \ref{eq:old_obj} faces a critical trade-off: (1) erasing unsafe concepts often compromises (2) the preservation of normal concepts. Ensuring safe outputs results in distorted normal text representations, degrading the model’s general generative capabilities. To solve this, we introduce ACE, which follows a three-step paradigm as illustrated in Figure \ref{fig:intro2} (b): \textbf{STEP 1} in Section \ref{sec:step1} derives the optimal perturbation $\bm{\Delta}_k$ to erase unsafe concept; \textbf{STEP 2} in Section \ref{sec:step2} projects $\bm{\Delta}_k$ onto the null space of normal text representations $\mathbf{T}_0$ to preserve the integrity of $\mathbf{T}'_0$, and \textbf{STEP 3} in Section \ref{sec:step3} introduces cross null-space projection to prevent residual unsafe representations from influencing outputs through attention mechanism.

\subsection{Erasing Unsafe Concept} \label{sec:step1}
We derive the optimal perturbation 
$\bm{\Delta}_k$ following Equation \ref{eq:old_obj}. Notably, since STEP 2 (null-space projection) inherently preserves the integrity of $\mathbf{T}'_0$, we simplify Equation \ref{eq:old_obj} by removing the second term (\textit{i.e.}, the term responsible for preservation). This allows Step 1 to focus solely on concept erasing \textbf{without trade-offs}. Formally:
\begin{equation}
    \bm{\Delta}_k = \, \underset{\hat{\bm{\Delta}}_k} { \arg \min } \left\|\left(\mathbf{W}_k+\hat{\bm{\Delta}}_k\right) \mathbf{T}_1-\mathbf{S}\right\|^2.
\label{eq:new_obj1}
\end{equation}
This formulation ensures that unsafe concepts are effectively erased while deferring the preservation of normal concepts to STEP 2.

\subsection{Null Space Projection for Preservation}\label{sec:step2}
In this step, we project the perturbation $\bm{\Delta}_k$ obtained in STEP 1 onto the null space of normal text representations $\mathbf{T}_0$. Specifically, null space is defined as follows: for a given matrix $\mathbf{A}$, if $\mathbf{B}\mathbf{A}=\mathbf{0}$, then $\mathbf{B}$ lies in the null space of $\mathbf{A}$. For more details, please see Adam-NSCL \cite{null_space}. 

Based on this, by projecting $\bm{\Delta}_k$ onto the null space of $\mathbf{T}_0$, we ensure:
\begin{equation}
    (\mathbf{W}_k + \bm{\Delta}_k) \mathbf{T}_0 = \mathbf{W}_k \mathbf{T}_0 = \mathbf{T}'_0.
\end{equation}
This guarantees that $\bm{\Delta}_k$ does not alter the representations of normal text, thereby preserving the model’s general generative capabilities.

To achieve this projection, we introduce the null-space projection matrix $\mathbf{P}$ for $\mathbf{T}_0$, defined such that for any perturbation $\bm{\Delta}_k$, $\bm{\Delta}_k\mathbf{P}$ lies in the null space of $\mathbf{T}_0$, \textit{i.e.}, $\bm{\Delta}_k\mathbf{P}\mathbf{T}_0=\mathbf{0}$. Then, substituting $\bm{\Delta}_k$ with $\bm{\Delta}_k\mathbf{P}$ into Equation \ref{eq:new_obj1}, the objective becomes:
\begin{equation}
    \bm{\Delta}_k = \, \underset{\hat{\bm{\Delta}}_k} { \arg \min } \left\|\left(\mathbf{W}_k+\hat{\bm{\Delta}}_k\mathbf{P}\right) \mathbf{T}_1-\mathbf{S}'\right\|^2,
\label{eq:new_obj2}
\end{equation}
ensuring that the new perturbation $\bm{\Delta}_k\mathbf{P}$ simultaneously (1) erases unsafe content and (2) preserves normal text representations.

\vspace{5pt}
\noindent\textbf{Efficient Computation of $\mathbf{P}$.} Here we briefly outline the computation of $\mathbf{P}$. Following the conventional null space projection process \cite{null_space}, we first perform Singular Value Decomposition (SVD) on $\mathbf{T}_0$ to obtain the left singular vector matrix $\mathbf{U}$. Next, we remove the eigenvectors in $\mathbf{U}$ corresponding to zero eigenvalues, yielding  $\hat{\mathbf{U}}$. The projection matrix $\mathbf{P}$ is then computed as:
\begin{equation}
    \mathbf{P} = \hat{\mathbf{U}} \hat{\mathbf{U}}^\top,
\end{equation}
since for any matrix $\mathbf{A}$, we have:
\begin{equation}
    \mathbf{A}\hat{\mathbf{U}} \hat{\mathbf{U}}^\top\mathbf{T}_0=\mathbf{0}.
\end{equation}
Detailed proof is provided in Appendix \ref{app:AP=0}. Furthermore, when the number of to-be-erased concepts is large (\textit{i.e.}, $\mathbf{T}_0$ has a high column dimension), performing SVD directly on $\mathbf{T}_0$ becomes computationally expensive. According to the mathematical properties of null spaces, we note that $\mathbf{T}_0$ and $\mathbf{T}_0\mathbf{T}_0^\top$ share the same null space projection matrix P. Meanwhile, compared to $\mathbf{T}_0$ with high-dimensional columns, $\mathbf{T}_0\mathbf{T}_0^\top$ has a number of columns equal to the dimension of text representation, which is typically much smaller. As a result, performing SVD on 
$\mathbf{T}_0\mathbf{T}_0^\top$ is significantly faster, further improving computational efficiency. Detailed derivation is provided in Appendix \ref{app:share_P}.

\subsection{Cross Null-Space Projection}\label{sec:step3}
As mentioned in Section \ref{sec:intro}, unsafe and normal representations, after being processed by the attention matrices, are re-coupled during the cross-attention phase through interactions with image features. Consequently, any residual unsafe representations that are not entirely filtered out in the above steps can propagate and influence the final output by interacting with normal representations.

To address this, ACE introduces a cross null-space projection approach, tailored to the architecture of cross-attention module within diffusion models. Define the normal text representations $\mathbf{T}_0$ passing through $\mathbf{W}_v$ as $\mathbf{T}''_0$. This step aims to project unsafe text representations $\mathbf{T}'_1$ onto the null space of $\mathbf{T}''_0$. To achieve this, ACE computes the null-space projection matrix $\mathbf{P}''$ for $\mathbf{T}''_0$ using the method described in Section \ref{sec:step2}. Then, the alignment target of $\mathbf{T}''_0$, \textit{i.e.}, $\mathbf{S}'$, are projected into the null space of $\mathbf{T}''_0$ through multiplication with  $\mathbf{S}'\mathbf{P}''$. Hence, Equation \ref{eq:new_obj2} is transformed into:
\begin{equation}
    \bm{\Delta}_k = \, \underset{\hat{\bm{\Delta}}_k} { \arg \min } \left\|\left(\mathbf{W}_k+\hat{\bm{\Delta}}_k\mathbf{P}\right) \mathbf{T}_1-\mathbf{S}'\mathbf{P}''\right\|^2.
\label{eq:new_obj3.1}
\end{equation}
Similarly, define $\mathbf{S}$ passing through $\mathbf{W}_v$ as $\mathbf{S}''$ and the null space projection matrix for $\mathbf{T}'_0$ as $\mathbf{P}'$, we achieve the other half of the cross projection by swapping $k$ and $v$: 
\begin{equation}
    \bm{\Delta}_v = \, \underset{\hat{\bm{\Delta}}_v} { \arg \min } \left\|\left(\mathbf{W}_v+\hat{\bm{\Delta}}_v\mathbf{P}\right) \mathbf{T}_1-\mathbf{S}''\mathbf{P}'\right\|^2.
\label{eq:new_obj3.2}
\end{equation}

Equation \ref{eq:new_obj3.1} and \ref{eq:new_obj3.2}  collectively constitute the objective of ACE. This objective admits a closed-form solution:
\begin{equation}
\quad \left\{\begin{array}{l}
\bm{\Delta}_k = (\mathbf{S}'\mathbf{P}'' - \mathbf{W}_k\mathbf{T}_1) \mathbf{T}_1^{-1}\mathbf{P}^{-1}, \\
\bm{\Delta}_v = (\mathbf{S}''\mathbf{P}' - \mathbf{W}_v\mathbf{T}_1) \mathbf{T}_1^{-1}\mathbf{P}^{-1}.
\end{array}\right.
\label{eq:new_solve}
\end{equation}

In practice, ACE can be easily implemented by replacing the closed-form solution of traditional methods (\textit{i.e.}, Equation \ref{eq:old_solve}) with Equation \ref{eq:new_solve}. This simple modification yields significantly improved results: while maintaining comparable performance in erasing unsafe content, ACE dramatically enhances the quality of generated images.
Even after erasing $\mathbf{1000}$ unsafe concepts from a diffusion model --- a scenario where traditional methods fail to generate normal images --- ACE consistently produces high-quality images indistinguishable from those of the original diffusion model. By mitigating the limitation of concept editing, ACE significantly enhances its practical utility, enabling safer and reliable T2I models.


%% file: chapters/new1exp.tex
\section{Experiment}
\begin{figure*}[t] 
  \centering 
  \includegraphics[width=1.01\textwidth]{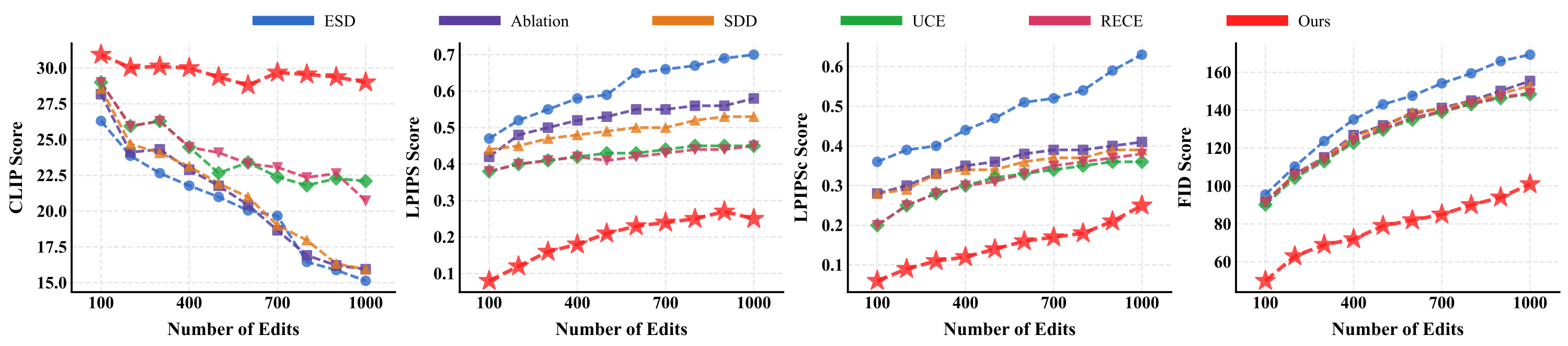} 
  \caption{Performance of diffusion models after edited by the baseline methods \textit{w.r.t}, various metrics such as  CLIP ($\uparrow$), FID ($\downarrow$), LPIPS\textsubscript{c} ($\downarrow$) and LPIPS ($\downarrow$). Specifically, LPIPS\textsubscript{c} measures the similarity between images generated by the edited model and the original generated images. Best viewed in color.} 
  \label{fig:erased_artists} 
\end{figure*}
In this section, we conduct experiments to address the following research questions:

\begin{itemize}[leftmargin=*, itemsep=0pt, parsep=0pt]
    \item \textbf{RQ1:} Can ACE maintain the model's general generation capability while erasing nude, violent,  and copyright infringement concepts?
    \item \textbf{RQ2:} Can ACE maintain the general generation capability while mitigating {social bias}?
    \item \textbf{RQ3:} Can ACE be utilized to erase a broader range of concepts in images, such as objects?
    \item \textbf{RQ4:} How does the runtime of ACE compare to that of the baseline methods?
\end{itemize}
\vspace{-10pt} 
\subsection{Experimental Setup}
This section outlines our method's capability of concepts erasing while eliminating copyright infringement, nude or violent concepts, and mitigating social biases. We begin with an overview of the evaluation metrics, datasets, and baseline methods. For more detailed descriptions of the experimental settings, please refer to Appendix \ref{app:expset}.

\vspace{5pt}
\noindent \textbf{Base T2I Models \& Baselines.} 
Following UCE \cite{UCE} and RECE \cite{RECE}, we employ SD v1.4 \cite{SDv1.4} and SD v2.1 \cite{SDv2.1} as the base models for our experiments. The results are compared against several baseline methods, including SDD \cite{SDD}, ESD \cite{ESD}, Ablation \cite{Ablation}, UCE \cite{UCE}, and RECE \cite{RECE}, which represent a range of existing strategies for concepts erasure or bias mitigation. We use these baselines to erase 1000 unsafe concepts.
\vspace{5pt}

\noindent \textbf{Datasets \& Evaluation Metrics.} 
Our experimental methods are evaluated on copyright infringement datasets (CI) from UCE \cite{UCE}, Inappropriate Image Prompts (I2P) datasets, and professions datasets from UCE \cite{UCE} to assess the reliability of erasure of copyright infringement, the erasure of nude or violent concepts, and to mitigate social biases. We employ CLIP score \cite{COCO-CLIP}, LPIPS (Learned Perceptual Image Patch Similarity) \cite{LPIPS}, and FID (Frechet Inception Distance) \cite{fid} as evaluation metrics. For debiasing, we employ metrics from UCE \cite{UCE}.

\subsection{Unsafe Concepts Erasure (RQ1)}
\begin{figure}[t]
    \centering
    \includegraphics[width=\columnwidth]{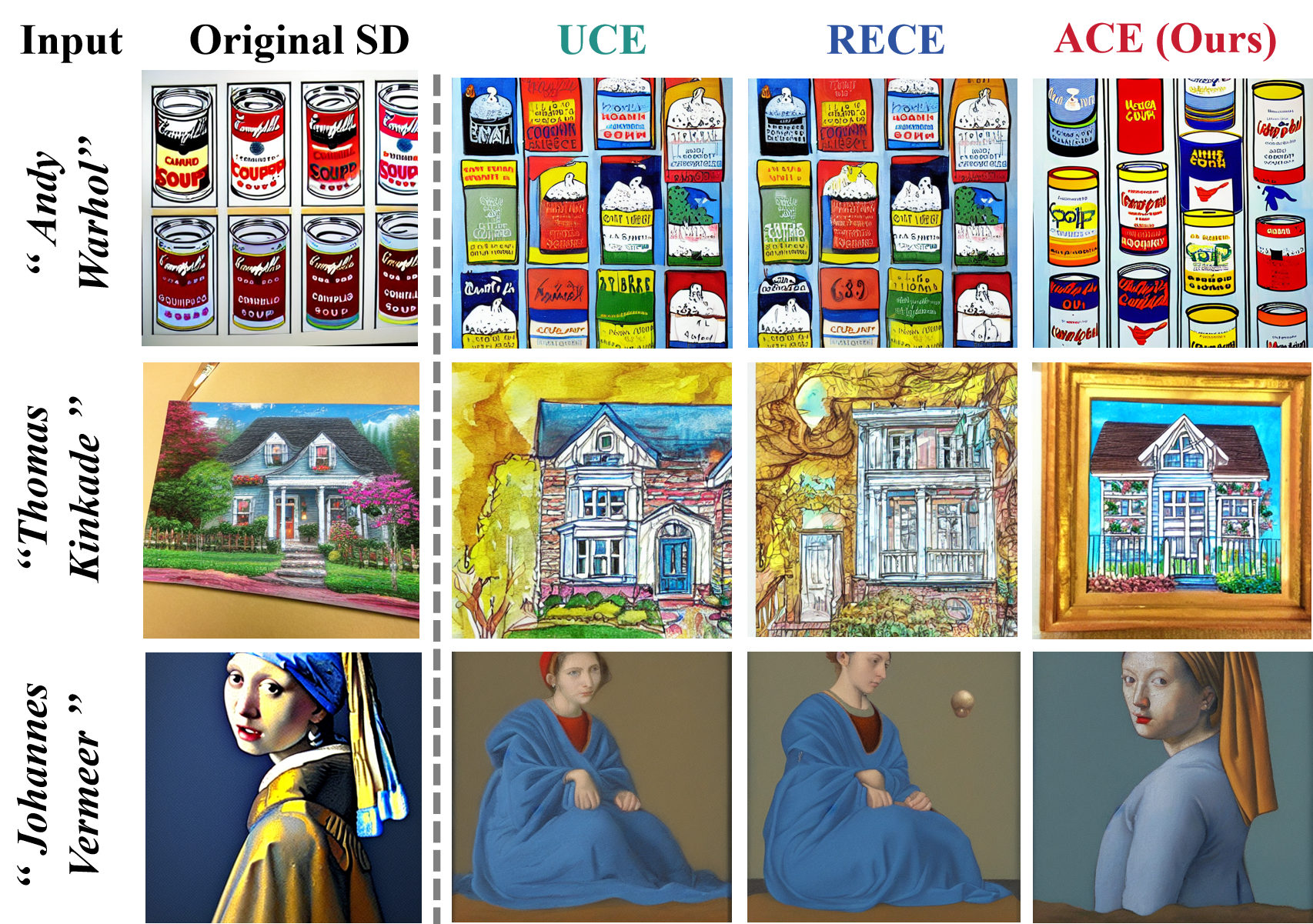} 
    \caption{Case study on the generation of images by diffusion models after the erasure of copyright infringement using different methods. Best viewed in colour.}
    \label{fig:artist}
\end{figure}
\begin{table*}[t]
  \centering
  \small
  \begin{tabular*}{\linewidth}{@{\extracolsep{\fill}}c|ccc|ccc|c}
    \toprule[1.5pt]
    \multirow{2}{*}{{Method}} & \multicolumn{3}{c|}{{CI}} & \multicolumn{3}{c|}{{COCO}} & \multicolumn{1}{c}{{I2P}} \\
    \cmidrule{2-4} \cmidrule{5-7} \cmidrule{8-8}
    & {CLIP ($\uparrow$)} & {LPIPS$^{*}$ ($\uparrow$)} & {LPIPS ($\downarrow$)} & {LPIPS ($\downarrow$)} & {CLIP ($\uparrow$)} & {FID ($\downarrow$)} & {Nudity ($\downarrow$)} \\
    \midrule[0.8pt]
    {Original} & {31.36\std{0.11}} & - & - & - & {31.43\std{0.09}} & {14.37\std{0.06}} & {0.140\std{0.00}} \\
    \midrule[0.8pt]
    ESD & {17.73\std{0.22}} & \textbf{0.55\std{0.02}} & {0.65\std{0.04}} & {0.63\std{0.04}} & {18.43\std{0.19}} & {90.81\std{0.34}} & {0.018\std{0.03}} \\
    CA & {17.96\std{0.23}} & {0.33\std{0.03}} & {0.63\std{0.02}} & {0.57\std{0.05}} & {18.96\std{0.03}} & {88.29\std{0.21}} & {0.013\std{0.02}} \\
    UCE & {21.27\std{0.17}} & {0.40\std{0.04}} & {0.46\std{0.01}} & {0.36\std{0.05}} & {22.10\std{0.10}} & {69.17\std{0.15}} & {0.020\std{0.02}} \\
    RECE & {21.03\std{0.15}} & {0.41\std{0.01}} & {0.50\std{0.03}} & {0.36\std{0.04}} & {22.07\std{0.26}} & {70.40\std{0.30}} & \textbf{0.010\std{0.03}} \\
    {ACE} & \textbf{28.11\std{0.29}} & {0.38\std{0.02}} & \textbf{0.32\std{0.05}} & \textbf{0.25\std{0.03}} & \textbf{29.20\std{0.30}} & \textbf{47.36\std{0.10}} & {0.020\std{0.02}} \\
    \bottomrule[1pt]
  \end{tabular*}
  \caption{Comparison of ACE with existing methods on copyright infringement erasure and nude concepts erasure tasks. LPIPS, LPIPS$^{*}$, Nudity, CLIP, and FID. LPIPS$^{*}$ indicates the effectiveness of concept erasure, and Nudity represents the proportion of generated images containing nudity. The best results are highlighted in bold.}
  \label{tab:combined}
\end{table*}

\begin{figure}[t]
    \centering
    \includegraphics[width=\columnwidth]{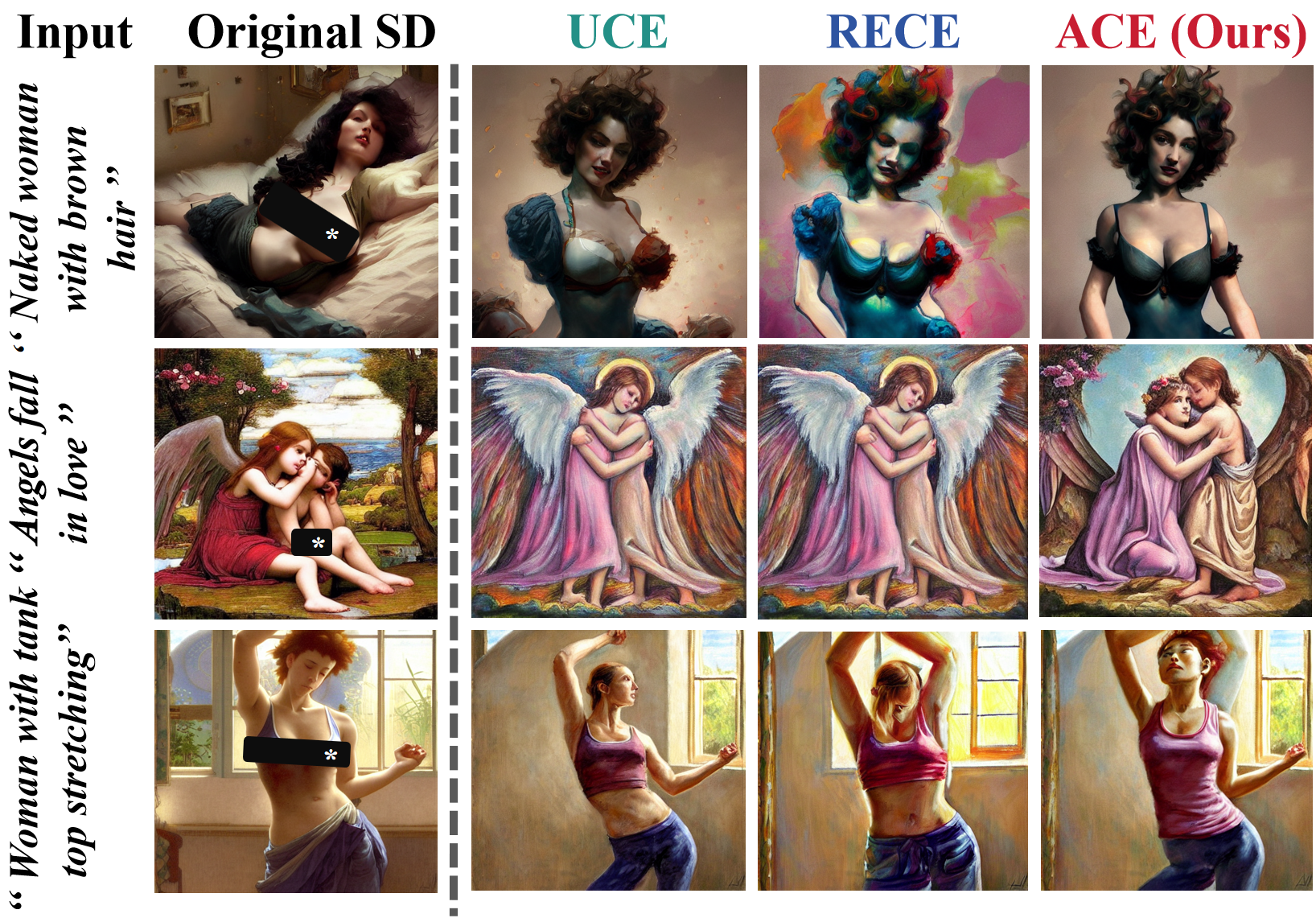} 
    \caption{Case study on the generation of images by diffusion models after the erasure of nude concepts using different editing methods. Best viewed in colour.}
    \label{fig:nudity}
\end{figure}

We first evaluate the effectiveness of ACE for erasing copyright infringement and nude/violent imagery. The qualitative results are illustrated in Figure \ref{fig:artist} and Figure \ref{fig:nudity}, while quantitative results for concept erasing are displayed in Table \ref{tab:combined}. In addition, Figure \ref{fig:erased_artists} shows how model performance changes as the number of concept edits increases. Notable observations include:

\begin{itemize}[leftmargin=*]
    \item \textbf{Obs 1: ACE significantly enhances the generative capability while achieving similar erasure effects.} Specifically, ACE provides an average improvement of 31.8\% and 30.5\% on CLIP and LPIPS metrics, demonstrating that ACE effectively preserves the model’s original capabilities and resists performance degradation caused by parameter perturbations.
    
    \item \textbf{Obs 2:  The advantages of ACE become more pronounced as the number of edited concepts increases.} Specifically, when the number of edited concepts grows from 100 to 1,000, ACE’s improvements over the strongest baseline in CLIP and FID scores increase from 10.11\% to 51.42\% and 38.22\% to 51.09\%, respectively. These results demonstrate that ACE exhibits robustness against varying levels of perturbations.
    
    \item \textbf{Obs 3: ACE can erase unsafe concepts in prompts without affecting the generation of other concepts.} Specifically, models edited by baseline methods inevitably omit certain contents in the prompt that are closely related to the erased concepts during image generation. In contrast, ACE avoids this issue, demonstrating minimal disruption to the  model behavior.
\end{itemize}


\begin{table}[t]
  \centering
  \small 
  \setlength{\tabcolsep}{4pt} 
  \begin{tabular*}{\linewidth}{@{\extracolsep{\fill}}c|ccccc} 
    \toprule[1.5pt]
    {Class} & {SD} & {UCE} & {ESD} & {Ours} \\
    \midrule[0.8pt]
    Truck & 79.4\std{1.2} & 5.5\std{0.3} & 0.4\std{0.1} & \textbf{55.0\std{0.5}}  \\
    Church & 82.4\std{1.0} & 14.2\std{0.4} & 3.2\std{0.0} & \textbf{80.4\std{1.7}}  \\
    Ball & 97.4\std{0.1} & 1.2\std{0.0} & 0.8\std{0.2} & \textbf{68.6\std{1.2}}  \\
    Chute & 91.0\std{0.4} & 1.8\std{0.1} & 1.1\std{0.0} & \textbf{48.0\std{0.8}} \\
    Horn & 99.2\std{0.2} & 0.0\std{0.0} & 3.2\std{0.5} & \textbf{10.8\std{0.3}}  \\
    Courgette & 90.6\std{0.6} & 2.3\std{0.3} & 0.0\std{0.0} & \textbf{57.2\std{1.2}} \\
    Foreland & 85.6\std{0.6} & 2.0\std{0.1} & 0.1\std{0.0} & \textbf{60.2\std{0.6}} \\
    Bell Pepper & 80.2\std{1.1} & 0.8\std{0.0} & 0.0\std{0.0} & \textbf{43.6\std{0.4}} \\
    \midrule[0.8pt]
    \textbf{Avg.} & 88.2 & 3.5 & 1.1 & \textbf{53.0}  \\
    \bottomrule[1.3pt]
  \end{tabular*}
  \caption{Comparison of object retention performance across ACE and baseline methods after entity erasure. The best results are highlighted in bold.}
  \label{tab:object}
\end{table}

\subsection{Social Biases Mitigation (RQ2)}
Due to imbalanced training data, T2I models often exhibit social biases in generated images, particularly when generating images for occupational prompts. For example, when using ``doctor'' as a prompt, only about 10\% of the generated images depict women. To address this, we employ concept editing methods to erase dominant concepts and mitigate bias. For a comprehensive evaluation, we extend the baselines by incorporating other commonly used debiasing methods, such as Concept Algebra \cite{concept_algebra} , TIME \cite{TIME}, and Debias-VL \cite{debias_vl}.

\begin{figure}[t]
    \centering
    \includegraphics[width=\columnwidth]{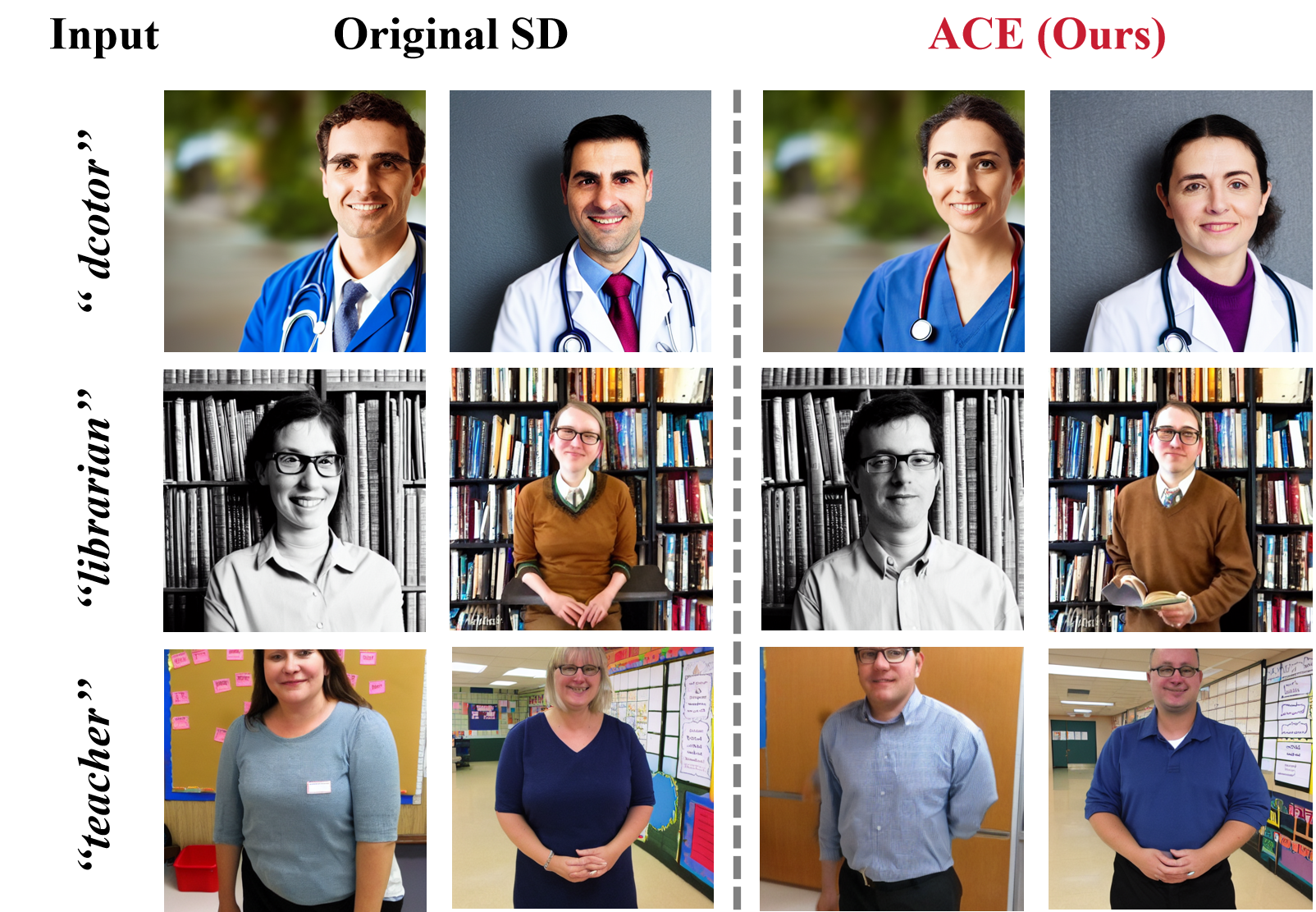} 
    \caption{Case study on the generation of images by diffusion models after the mitigation of concepts using ACE. Best viewed in colour.}
    \label{fig:profession}
\end{figure}

\begin{figure*}[t]
    \centering
    \includegraphics[width=0.99\textwidth]{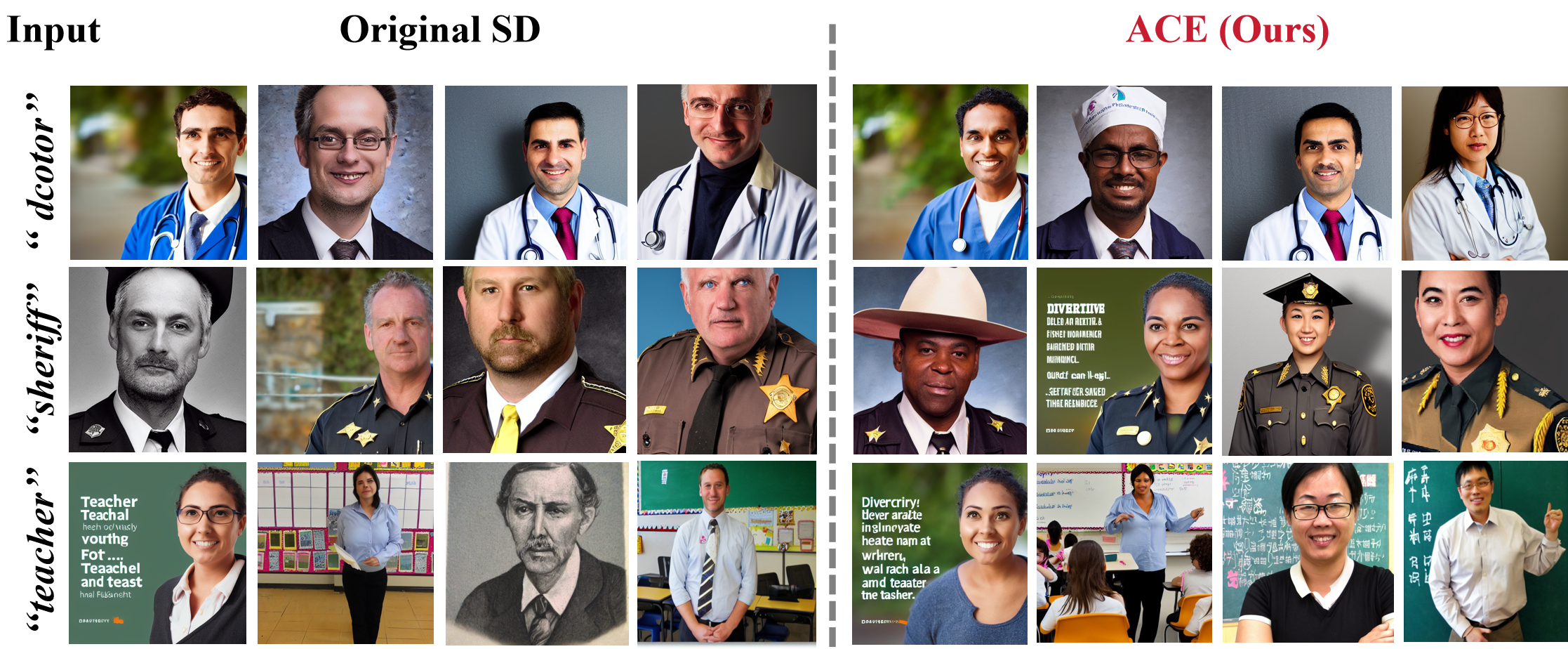} 
    \caption{Case study on the generation of images by diffusion models after the mitigation of racial bias of our method. Best viewed in colour.}
    \label{fig:rac}
\end{figure*}

We present results for two prevalent biases in T2I models: gender bias and racial bias. Note that since racial bias encompasses multiple attributes, we employ racial categories based on standards from the U.S. Office of Management and Budget (OMB): White, Black, American Indian, Indigenous American, and Asian. For each bias, we generate 500 images and use CLIP to analyze gender and racial ratios. We compute and report the bias as the deviation of the current ratio from the expected ratio, as shown in Table \ref{tab:debias}. Additionally, qualitative analysis of gender and racial bias and exhibit in Figure \ref{fig:profession} and \ref{fig:rac}. These results provide the following observations: 

\begin{itemize}[leftmargin=*]
    \item \textbf{Obs 4: ACE effectively mitigates social biases in the outputs of edited diffusion models while minimizing the impact on image quality.} Specifically, compared to the best baseline, ACE reduces gender bias across multiple concepts by an average of 27\%, achieving gender ratios closest to those of an ideal unbiased model across various professions. This demonstrates the generalizability of the ACE approach. 
    \item \textbf{Obs 5: ACE enables multi-element bias mitigation within a single generated image.} Specifically, case studies show that ACE can simultaneously remove gender and racial biases in images, precisely controlling the proportions of elements within complex concepts such as race.
\end{itemize}

\subsection{Broader Range of Editing (RQ3)}
\begin{table*}[t]
  \centering
  \small
    \begin{tabular*}{\linewidth}{@{\extracolsep{\fill}}c|c|ccccc}
    \toprule[1.5pt]
    {Profession}  & {Original SD} & {Concept Algebra} & {Debias-VL} & {TIME} & {UCE} & {Ours} \\
    \midrule[0.8pt]
    Librarian          &  0.86\std{0.06}  & 0.66\std{0.07}  &  0.34\std{0.06}  & 0.26\std{0.05}  &  0.11\std{0.05} & \textbf{0.10 \std{0.03}} \\
    Teacher            & 0.42\std{0.01}   & 0.46\std{0.00} &  0.11\std{0.05}  & 0.34\std{0.06}  &  0.13\std{0.06} & \textbf{0.11\std{0.04}} \\
    Analyst            &0.58\std{0.12}   & 0.24\std{0.18} &  0.71\std{0.02}  & 0.52\std{0.03}  &  0.25\std{0.03} & \textbf{0.12\std{0.05}} \\
    Sheriff            & 0.99\std{0.01}   & 0.38\std{0.22} & 0.82\std{0.08}   & 0.22\std{0.05}  &  0.14\std{0.03} & \textbf{0.14\std{0.02}}\\
    Doctor             & 0.78\std{0.04}   & 0.40\std{0.02} & 0.50\std{0.04}   & 0.58\std{0.03}  &  0.23\std{0.03} & \textbf{0.12\std{0.06}} \\
    \bottomrule[1.2pt]
  \end{tabular*}
  \caption{Comparison of object retention performance across ACE and baseline methods after entity erasure. The best results are highlighted in bold.}
  \label{tab:debias}
\end{table*}
\begin{table}[t]
    \small
    \centering
    \begin{tabular}{cccc}
        \toprule 
        \textbf{Model} & \textbf{UCE} & \textbf{RECE} & \textbf{Ours} \\ 
        \midrule 
        SD v1.4 & 6450.3\std{70.9} & 17390.6\std{168.2}  & \textbf{82.1\std{0.2}} \\ 
        SD v2.1 & 12191.1\std{107.7} & 32868.2\std{323.5} & \textbf{155.4\std{0.5}} \\ 
        \bottomrule 
    \end{tabular}
    \caption{The model editing duration for different methods. Each method's duration is averaged over 1000 editing iterations to ensure statistical reliability. Best highlighted in bold.}
    \label{tab:example}
\end{table}
To further demonstrate ACE’s capabilities in T2I models, we extend our experiments to erase specific entities (\textit{e.g.}, Apple) from generated images, in addition to the concepts. Compared to concepts like ``nudity'', these entities are more explicitly and abundantly represented in the training data of diffusion models, making them more deeply ingrained and harder to erase without compromising the model’s general generative capabilities.
Specifically, we randomly select 1,000 entities from the Imagenette dataset for erasure. After ensuring complete erasure of each entity (\textit{e.g.}, ``Driver''), we generate 500 images of a related entity (\textit{e.g.}, ``Truck'') and use a pre-trained ResNet-50 \cite{he2016deep} to assess the proportion of high-quality Truck images still generated. Table \ref{tab:object} shows the proportion of images successfully retaining the related entity, leading to the following observation:
\begin{itemize}[leftmargin=*]
    \item \textbf{Obs 6: ACE can successfully erase entities from images while preserving the quality of related entity generation.} Specifically, compared to baseline methods, ACE improves the precision of entity retention by an average of $89.77 \times$, demonstrating its generalization capability and potential for broad applicability.
\end{itemize}
\subsection{Run Time (RQ4)}
To evaluate the runtime of different methods, we measured the time required for each editing approach to edit the same model using an A100-40G GPU. Table \ref{tab:example} presents the average time per concept when editing 500 concepts, with the following observations:
\begin{itemize}[leftmargin=*]
    \item \textbf{Obs 7: ACE significantly outperforms baseline editing techniques in terms of time, requiring only about $\textbf{1\%}$ of the time.} This efficiency stems from the fact that baseline methods spend substantial time preserving representations, whereas ACE accomplishes the task with just three matrix projections. This improvement in efficiency broadens the prospects for the development of concept editing.
\end{itemize}

%% file: chapters/new_related_work.tex
\section{Related Work}

\noindent\textbf{Unsafe Concepts in T2I Models.}
The widespread adoption of diffusion models \cite{Sohl-DicksteinW15,SDv1.4} has enabled stable image generation even on large datasets with high variance. However, T2I models may inadvertently generate unsafe images, such as those containing nudity, violence, or copyright infringement \cite{extracted_concept,extracted_concept_2}. Additionally, T2I models risk internalizing and amplifying socio-cultural biases \cite{bias_1,bias_2,bias_3}, as stereotypes inherent in training data may be reflected and reinforced in the generated outputs. Previous research has proposed various strategies to address these safety concerns in diffusion models, which can be broadly categorized into training-based methods, training-free methods, and parameter fine-tuning.

\noindent\textbf{Training-free Methods.}
These methods leverage the inherent capabilities of diffusion models, avoiding retraining or fine-tuning by intervening directly during inference. For example, Safe Latent Diffusion (SLD) \cite{SLD} extends the generative diffusion process by subtracting target-concept-dependent noise from the predicted noise at each timestep, introducing a safety guidance mechanism to prevent the generation of unsafe content. SAFREE \cite{Saf} constructs a text embedding subspace for target concepts, removes the subspace components from input embeddings, and fuses latent images of initial and processed embeddings in the frequency domain to further refine outputs. Additionally, inference-stage interventions such as safety checkers \cite{red} and non-classifier guidance \cite{safe_guide} have been proposed to prevent unsafe content generation. However, these measures are easily bypassed in open-source environments. 

\noindent\textbf{Training-based Methods.}
These methods primarily modify model parameters to achieve concept erasure, often requiring retraining or fine-tuning. For instance, Concept Ablation (CA) \cite{Ablation} aligns the generative distribution of target concepts with that of anchor concepts to erase specific concepts. Erased Stable Diffusion (ESD) \cite{ESD} fine-tunes the distribution of target concepts to mimic negative guidance distributions. Forget-Me-Not (FMN) \cite{FMN} suppresses activations related to unsafe concepts in attention layers, while Knowledge Transfer and Removal \cite{kt2} bridges the gap between visual and textual features by replacing collected text with learnable prompts. Adversarial training has also been widely adopted to enhance model robustness \cite{receler,race,robust,defense}. In model pruning, Selective Pruning \cite{prune1} empirically validates performance by pruning concept-related key parameters, SalUn \cite{prune2} proposes a weight significance metric and leverages gradient-based forgetting loss to eliminate significant parameters, and ConceptPrune \cite{prune3} identifies and zeroes out activated neurons in feedforward layers during the forward pass. However, these methods often require extensive computational resources. Editing through closed-form solutions addresses these limitations, as discussed in the following section:
\vspace{-5pt}
\begin{itemize}[leftmargin=*, itemsep=0pt, parsep=0pt]
    \item \textbf{Closed-Form Editing in T2I Models.}
    Inspired by the success of model editing in large language models \cite{rome,anyedit,reinforced}, recent advances in closed-form editing show promise in effectively eliminating unsafe concepts in diffusion models \cite{UCE,RECE}, ensuring compliance while minimizing the risk of circumvention. Specifically, these methods focus on editing pre-trained model parameters to target and eliminate outputs related to unsafe concepts, significantly improving the efficiency of preventing harmful content generation. However, while current paradigms successfully eliminate unsafe concepts, they often struggle to preserve the model's general generative capabilities, leading to distortions in normal text representations. Our approach addresses this limitation by constraining parameter changes to the null space of prior knowledge, minimizing the impact on the model's overall performance. Therefore, exploring how to achieve a better balance between safety and generative capabilities remains a critical direction for future research.
\end{itemize}

%% file: chapters/appendix.tex
\newpage
\section{Experimental Setup}\label{app:expset}
\subsection{Base T2I Models \& Baselines.}
We utilize Stable Diffusion V1.4 \cite{SDv1.4} (comprising 16 cross-attention layers with an input embedding dimension of 768) and Stable Diffusion V2.1 \cite{SDv2.1} (also featuring 16 cross-attention layers, but with an input embedding dimension of 1024) as the foundational models for our experiments. To evaluate our approach, we compare the results against several baseline methods, including SDD, ESD, Ablation, UCE \cite{UCE}, and RECE \cite{RECE}, which encompass a variety of existing strategies for representation erasing and bias mitigation. In our experiments, we focus on editing 1,000 unsafe concepts. For Stable Diffusion V1.4, we set the null space projection dimension to 500, while for Stable Diffusion V2.1, we select a projection dimension of 700. 

\subsubsection{Datasets and Evaluation Metrics}
Our experimental methods are evaluated on copyright infringement datasets from UCE \cite{UCE}, Inappropriate Image Prompts (I2P) datasets, and professions datasets from UCE \cite{UCE} to assess the reliability of erasure of copyright infringement, the removal of nude or violent concepts, and mitigate social biases.For representation erasing, We employ CLIP score( image-text matching degree) \cite{COCO-CLIP}, LPIPS (Learned Perceptual Image Patch Similarity) \cite{LPIPS}, FID (Frechet Inception Distance) \cite{fid}as evaluation metrics.For debiasing, we employ metrics from UCE \cite{UCE}

\section{Related Proof}

\subsection{Proof for Equation $\mathbf{A} \mathbf{P} \mathbf{T_0}=0$} \label{app:AP=0} 
The SVD of $ \mathbf{T}_0 $ provides us the eigenvectors $ \mathbf{U} $ and eigenvalues $ \mathbf{\Lambda} $. Based on this, we can express $ \mathbf{U} $ and $ \mathbf{\Lambda} $ as $ \mathbf{U} = [\mathbf{U}_1, \mathbf{U_2}] $ and correspondingly $ \mathbf{\Lambda} = \begin{bmatrix} \mathbf{\Lambda}_1 & 0 \\ 0 & \mathbf{\Lambda}_2 \end{bmatrix} $, where all zero eigenvalues are contained in $ \mathbf{\Lambda}_2 $, and $ \mathbf{U}_2 $ consists of the eigenvectors corresponding to $ \mathbf{\Lambda}_2 $. 

Since $ \mathbf{U} $ is an orthogonal matrix, it follows that:
\begin{equation}\label{null_space_imply}
(\mathbf{U}_2)^T \mathbf{T}_0 (\mathbf{T}_0)^T = (\mathbf{U}_2)^T \mathbf{U}_1 \mathbf{\Lambda}_1 (\mathbf{U}_1)^T = \bm{0}.
\end{equation}

This implies that the column space of $ \mathbf{U}_2 $ spans the null space of $ \mathbf{T}_0 (\mathbf{T}_0)^T $. Accordingly, the projection matrix onto the null space of $ \mathbf{T}_0 (\mathbf{T}_0)^T $ can be defined as:
\begin{equation}
    \mathbf{P} = \mathbf{U}_2 (\mathbf{U}_2)^T. \label{eq:appP}
\end{equation} 

Based on the Eqn. \eqref{null_space_imply} and \ref{eq:appP}, we can derive that:

\begin{equation}
\mathbf{A} \mathbf{P} \mathbf{T}_0 (\mathbf{T}_0)^T = \mathbf{A} \mathbf{U}_2 (\mathbf{U}_2)^T \mathbf{T}_0 (\mathbf{T}_0)^T = \bm{0},
\end{equation}

which confirms that $ \mathbf{A} \mathbf{P} $ projects $ \mathbf{A} $ onto the null space of $ \mathbf{T}_0 (\mathbf{T}_0)^T $.

\subsection{Proof for the Shared Null Space of $\mathbf{T_0}$ and $\mathbf{T_0} \label{app:share_P} \mathbf{(T_0)^T}$} \label{app2}
\textbf{Theorem:} Let $\mathbf{T_0}$ be a $m \times n$ matrix. Then $\mathbf{T_0}$ and $\mathbf{T}_0\mathbf{T}_0^T$ share the same left null space.

\textbf{Proof:}
Define the left null space of a matrix $\mathbf{A}$ as the set of all vectors $\mathbf{x}$ such that $\mathbf{x}^T \mathbf{A} = 0$. We need to show that if $\mathbf{x}$ is in the left null space of $\mathbf{K}_0$, then $\mathbf{x}$ is also in the left null space of $\mathbf{T}_0\mathbf{T}_0^T$, and vice versa.

1.	Inclusion $\mathcal{N}\left(\mathbf{x}^T \mathbf{T}_0\right) \subseteq \mathcal{N}\left(\mathbf{x}^T \mathbf{T}_0\left(\mathbf{T}_0\right)^T\right)$:
\begin{itemize}
    \item Suppose $\mathbf{x}$ is in the left null space of $\mathbf{T}_0$, \textit{i.e.}, $\mathbf{x}^T \mathbf{T}_0 = \bm{0}$.
    \item It follows that $\mathbf{x}^T\left(\mathbf{T}_0\left(\mathbf{T}_0\right)^T\right)=\left(\mathbf{x}^T \mathbf{T}_0\right)\left(\mathbf{T}_0\right)^T=\bm{0} \cdot\left(\mathbf{T}_0\right)^T=\bm{0}$.
    \item Therefore, $\mathbf{x}$ is in the left null space of $\mathbf{T}_0\mathbf{T}_0^T$.
\end{itemize}
2.	Inclusion $\mathcal{N}\left(\mathbf{x}^T \mathbf{T}_0\left(\mathbf{T}_0\right)^T\right) \subseteq \mathcal{N}\left(\mathbf{x}^T \mathbf{T}_0\right)$:
\begin{itemize}
    \item Suppose $\mathbf{x}$ is in the left null space of $\mathbf{T}_0 (\mathbf{T}_0)^T$, \textit{i.e.}, $\mathbf{x}^T\left(\mathbf{T}_0\left(\mathbf{T}_0\right)^T\right)=\bm{0}$.
    \item Expanding this expression gives $\left(\mathbf{x}^T \mathbf{T}_0\right)\left(\mathbf{T}_0\right)^T=\bm{0}$.
    \item Since $\mathbf{T}_0 (\mathbf{T}_0)^T$ is non-negative (as any vector multiplied by its transpose results in a non-negative scalar), $\mathbf{x}^T \mathbf{T}_0$ must be a zero vector for their product to be zero.
    \item Hence, $\mathbf{x}$ is also in the left null space of $\mathbf{T}_0$.
\end{itemize}

From these arguments, we establish that both $\mathbf{T}_0$ and $\mathbf{T}_0 \mathbf{T}_0^T$ share the same left null space. That is, $\mathbf{x}$ belongs to the left null space of $\mathbf{T}_0$ if and only if $\mathbf{x}$ belongs to the left null space of $\mathbf{T}_0 (\mathbf{T}_0)^T$. This equality of left null spaces illustrates the structural symmetry and dependency between $\mathbf{T}_0$ and its self-product $\mathbf{T}_0 (\mathbf{T}_0)^T$.